\documentclass[letterpaper, 10 pt, conference]{ieeeconf}  

\IEEEoverridecommandlockouts                              

\usepackage{epsfig}
\usepackage{makecell}
\usepackage{subcaption}
\usepackage{xcolor}
\usepackage{ragged2e}
\usepackage{float} 
\usepackage{graphicx}
\usepackage{amsmath}
\usepackage{adjustbox}
\usepackage{amssymb}
\usepackage{booktabs}
\usepackage{tabularx}
\usepackage{array}
\usepackage{makecell}
\usepackage{bm}

\title{Next-Generation Travel Demand Modeling with a Generative Framework for Household Activity Coordination}

\author{Xishun Liao$^{1}$, Haoxuan Ma$^{1}$, Yifan Liu$^{1}$, Yuxiang Wei$^{1}$, Brian Yueshuai He$^{2}$, Chris Stanford$^{3}$, and Jiaqi Ma*$^{1}$
\thanks{$^{1}$ UCLA Mobility Lab under the Department of Civil and Environmental Engineering, University of California, Los Angeles, Los Angeles, USA.}
\thanks{$^{2}$ University of Louisville, Louisville, KY, USA}
\thanks{$^{3}$ Novateur Research Solutions, Ashburn, VA, USA.}
\thanks{*Corresponding author: jiaqima@ucla.edu}
}

\begin{document}

\maketitle
\thispagestyle{empty}
\pagestyle{empty}

\begin{abstract}
Travel demand models are critical tools for planning, policy, and mobility system design. Traditional activity-based models (ABMs), although grounded in behavioral theories, often rely on simplified rules and assumptions, and are costly to develop and difficult to adapt across different regions. This paper presents a learning-based travel demand modeling framework that synthesizes household-coordinated daily activity patterns based on a household's socio-demographic profiles. The whole framework integrates population synthesis, coordinated activity generation, location assignment, and large-scale microscopic traffic simulation into a unified system. It is fully generative, data-driven, scalable, and transferable to other regions. A full-pipeline implementation is conducted in Los Angeles with a 10 million population. Comprehensive validation shows that the model closely replicates real-world mobility patterns and matches the performance of legacy ABMs with significantly reduced modeling cost and greater scalability. With respect to the SCAG ABM benchmark, the origin-destination matrix achieves a cosine similarity of 0.97, and the daily vehicle miles traveled (VMT) in the network yields a 0.006 Jensen-Shannon Divergence (JSD) and a 9.8\% mean absolute percentage error (MAPE). When compared to real-world observations from Caltrans PeMS, the evaluation on corridor-level traffic speed and volume reaches a 0.001 JSD and a 6.11\% MAPE.
\end{abstract}

\section{Introduction}
Accurate travel demand models support urban planning, mobility system optimization, commercial strategy, public health, and safety \cite{ma2024mobility,  pappalardo2016analytical, abidi2023mobility, bohm2021quantifying}, by providing detailed forecasts of where, when, how, and why people travel. Over time, travel demand modeling approaches have evolved from trip-based models \cite{ben1985discrete, horowitz1980utility}, which treated individual trips as independent units, to tour-based models \cite{algers1995stockholm, rossi1997tour} that grouped linked trips into daily tours, and ultimately to activity-based models \cite{goulias2011simulator, bowman2001activity, bhat2012household} that represent travel as a function of a person's activity agenda over a day. Despite their advancements, traditional activity-based travel demand models still face key limitations. They rely on simplified rules and assumptions that may not reflect the variety and complexity of real-world behavior. Building and applying these models requires extensive data, development, and calibration, making them costly and time-consuming. Their complexity also results in high computational demands, and built-in assumptions limit their adaptability across regions.

To address these challenges, our previous work introduced a deep generative activity model \cite{liao2024deep} trained on the National Household Travel Survey (NHTS), capable of generating daily activity chains and assigning locations based on socio-demographic profiles. The resulting activity and traffic patterns, such as speed, flow, and origin-destination distributions, closely match the observed data, and the model transfers well to other regions with minimal local data. However, it models individuals in isolation, overlooking the intra-household coordination seen in real-world travel behavior, including shared trips and joint schedules.

While prior research has explored household coordination through rule-based or utility-based methods, learning-based approaches to modeling coordinated activity patterns remain underdeveloped. This paper develops a transferable and scalable travel demand modeling framework, which extends our previous deep generative model for individuals to incorporate intra-household coordination. Given socio-demographic profiles of households, the model generates daily activity patterns and travel behavior. The proposed framework will capture the who, when, where, why, and how of daily travel decisions. Compared to existing literature, our key contributions include:

\begin{itemize}
    
    \item We developed a next-generation travel demand modeling framework that synthesizes travel behavior for city residents. This scalable and transferable framework enables automated transportation system simulation and large-scale human mobility data synthesis in new regions, advancing the state of the art in travel demand modeling.
    
    \item We introduced the Deep Coordinated Activity Model (DeepCAM) for household-coordinated activity synthesis. DeepCAM captures the interdependence among household members and their activities, enabling coordinated activity generation. 

    \item We proposed a next-generation travel demand modeling pipeline that integrates population synthesis, activity generation, and large-scale agent-based traffic simulation. This pipeline was demonstrated in a full-scale case study of Los Angeles (LA), with validation against real-world observations using comprehensive metrics for both individual behavior and intra-household coordination.
\end{itemize}

\section{Literature review}

\subsection{Activity-Based Travel Demand Model}
Activity-based models (ABMs) represent the state-of-the-art in travel demand modeling by simulating daily activity-travel behavior based on individuals' activity agendas. Bowman and Ben-Akiva \cite{bowman2001activity} introduced a discrete choice-based ABM that captures full-day activity patterns and travel decisions. Building on this, SimAGENT \cite{goulias2011simulator,bhat2012household} integrated population synthesis, land use, and travel simulation to model the Southern California region. These models improve behavioral realism but are costly to develop, data-intensive, and difficult to transfer across regions. Recent efforts address these challenges through more flexible and transparent modeling frameworks. SimMobility \cite{adnan2016simmobility}, developed for Singapore, integrates daily activity-based demand modeling with network and land-use simulations in a unified, agent-based platform, supporting complex, large-scale policy analysis. In parallel, the ActivitySim platform \cite{freedman2023activitysim} was created as an open-source, modular ABM focused on regional adaptation and collaborative development. It builds on established utility-based designs while emphasizing ease of deployment, transparency, and extensibility for planning agencies.


However, these models still rely on predefined choice structures with handcrafted assumptions. Learning-based models offer a more flexible alternative. Liao et al. \cite{liao2024deep} proposed a transformer-based model trained on survey data that generates activity-location chains given socio-demographic profiles and transfers across regions with minimal tuning. Similarly, Shone and Hillel \cite{shone2025modelling} adopted variational autoencoders to synthesize individual activity schedules based on the learned distribution of travel survey data. Although these learning-based methods enhance scalability and realism, they have largely focused on individuals and have not considered coordinated activity.

\subsection{Household Coordinated Activity Modeling}
Household coordination is essential in travel demand modeling, as many travel decisions involve joint activities and shared responsibilities. Gliebe and Koppelman \cite{gliebe2005modeling} analyzed weekday activity patterns of two-person households and proposed a utility-based structural model for coordinated daily schedules. Bradley and Vovsha \cite{bradley2005model} extended this with a rule-based utility framework for households of up to five members, capturing joint decision-making.

To capture heterogeneity in household decision processes, Zhang et al. \cite{zhang2009modeling} introduced a latent class model incorporating altruistic and dominant strategies, showing that employed males and primary car users often exert greater influence. Shakeel et al. \cite{shakeel2022joint} applied a similar model to weekly activity generation, identifying four behavioral household segments shaped by structure and schedules. While Zhang focused on intra-household dynamics, Shakeel emphasized inter-day variation in joint activity behavior.

Building on the need to capture more dynamic and interactive household behavior, Arentze and Timmermans \cite{arentze2009need} proposed a need-based model for multi-day, multi-person activity generation. In this framework, individuals select activities based on evolving needs and adaptive utility-of-time thresholds, while accounting for within-household interactions. Bhat et al. \cite{bhat2013household} later developed a comprehensive household-level model using a multiple discrete-continuous extreme value framework. It predicts both independent and joint activity participation across all household members and activity types. Estimated on Southern California survey data, the model integrates detailed attributes and offers computational efficiency, making it suitable for large-scale simulation.

While these aforementioned models are effective, they are limited by the assumptions about decision rules or household segmentation, lacking adaptability to varying contexts. They also face data constraints, with high costs in data collection, model development, and calibration, and pose significant computational demands. A learning-based modeling approach is thus needed to flexibly infer coordination behaviors directly from data without heavily relying on pre-defined assumptions.

\begin{figure}
  \centering
  \includegraphics[width=0.43\textwidth]{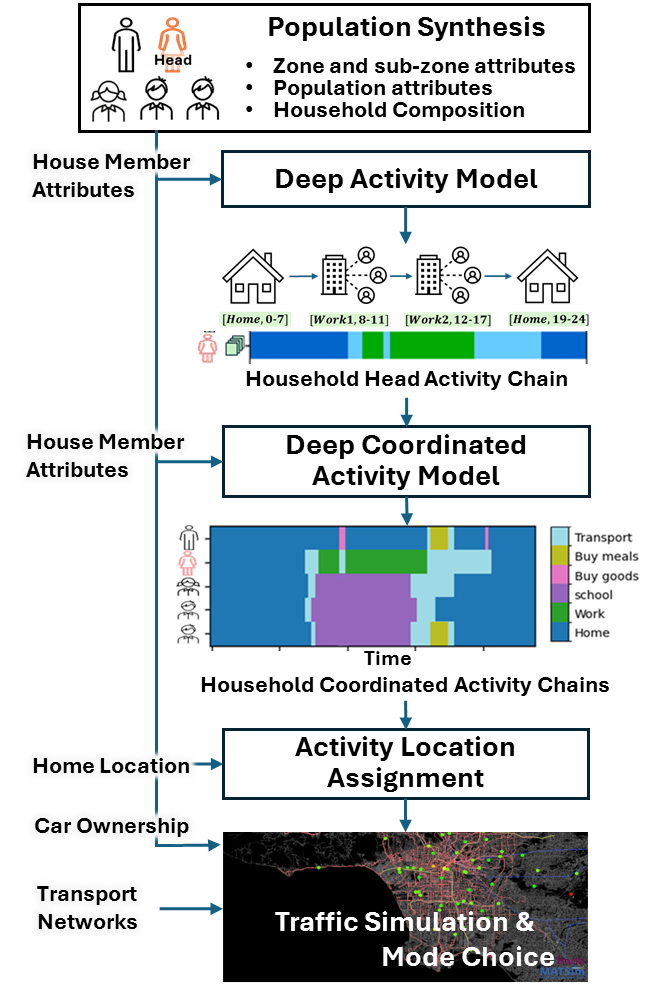}
  \caption{System framework of the proposed next generation travel demand model.}
  \label{fig:system}
\end{figure}

\section{Problem Formulation}
Based on the household profiles, the travel demand framework generates daily travel trajectories, including individual activities, household-level interactions, activity locations, route and mode choices, and the resulting network loads.

The proposed generative framework, illustrated in Fig.~\ref{fig:system}, begins with population synthesis, creating synthetic households based on census statistics. Each agent $p_i$ belongs to a household $H$ described by household-level features $\mathbf{h}$, and each agent is further characterized by individual socio-demographic features $\mathbf{f}_{p_i}$.

Then, given the $\mathbf{h}$ and $\mathbf{f}_{p_i}$, the Deep Activity Model (DeepAM) \cite{liao2024deep} generates an initial activity chain for the household head. An activity chain for agent $p_i$ is defined as a time-ordered sequence $A_{p_i} = \left\{A_{1,{p_i}}, A_{2,{p_i}}, \ldots, A_{n,{p_i}}\right\}$, where each element $A_{n,{p_i}} = \left[T_{n,{p_i}}, S_{n,{p_i}}, E_{n,{p_i}}\right]$ represents the $n$-th activity conducted by agent $p_i$. Here, $T_{n,{p_i}}$ denotes the activity type, while $S_{n,{p_i}}$ and $E_{n,{p_i}}$ represent the start time and end time of the activity, respectively.

The seed chain $A_{pi}$ of the household head, together with the household attributes $\mathbf{h}$ and the socio-demographic features of all household members $\left\{\mathbf{f}_{p_1}, \ldots, \mathbf{f}_{p_n}\right\}$, is used by the DeepCAM to generate activity chains $\left\{A_{p_1}, \ldots, A_{p_n}\right\}$ for other household members, capturing shared activities and inter-personal dependencies.

A comprehensive travel demand model assigns each activity in a chain with a zonal-level location $z_{i,{p_i}}$ where the $A_{i, {p_i}}$ occurs. Next, these location-annotated trajectories are then input into a simulation engine, where mode choice $m_{i, {p_i}}$ and route choice $r_{i, {p_i}}$ of each activity and traffic dynamics are subsequently evaluated. Finally, we obtain travel demand in trajectories for all residents of a city, as $Traj_{p_i} = \left\{ \left( A_{1, {p_i}}, z_{1,{p_i}}, m_{1, {p_i}}, r_{1, {p_i}} \right), \ldots, \left( A_{n, {p_i}}, z_{n,{p_i}}, m_{n, {p_i}}, r_{n, {p_i}} \right)  \right\}$.

\section{Methodology}
\subsection{Population Synthesis}
The population synthesis module, shown in Fig.~\ref{fig:system}, initializes the generative framework by creating a synthetic population aligned with regional census distributions. Each household is assigned attributes such as size, income, vehicle ownership, and home location at the TAZ level, while individual members receive characteristics such as age, employment status, student status, education, and license ownership. Households serve as the core unit, enabling coordinated activity generation across members in subsequent stages.

We adopt a modular design in which the population synthesis process is decoupled from activity generation and coordination modules. This enables flexibility in the choice of synthesis tools and ensures that the overall framework remains adaptable across geographies and modeling contexts. Standard population synthesis tools such as PopGen \cite{mcnally2014popgen,marg2016popgen} and SimAGENT \cite{pendyala2012simagent} can be readily integrated, and more recent data-driven or generative alternatives \cite{farooq2013simulation} can also be used without altering the downstream modeling logic.

\subsection{Activity Generation}
\subsubsection{Household Head Activity Generation}
DeepAM, developed in our previous work \cite{liao2024deep}, is a generative deep learning model designed to synthesize realistic daily activity chain for individuals. Trained on the NHTS, it captures the complex relationships between activity patterns and a wide range of inputs, including 13 types of socio-demographic attributes of the target individual and household members, as well as 13 types of household and zonal characteristics such as vehicle ownership, home ownership, population density, and residential classification.

DeepAM generates activity chain autoregressively for each activity $A_{i, {p_i}}$ until a special end-of-day token is emitted. Built on a transformer-based architecture, DeepAM effectively models temporal dependencies and contextual influences, capturing how past activities, as well as the presence and roles of other household members, inform the generation of future activities. 

Demonstrated high fidelity in synthesizing observed activity patterns across multiple regions. By generating individualized activity chains that closely align with empirical data, DeepAM serves as a foundational component for simulating travel demand. Building on this foundation, the DeepCAM extends DeepAM's capabilities to model household-level activity planning and coordination. Since decision-making within households often skews toward employed males and primary vehicle users \cite{zhang2009modeling}, the activity chain of the household head, generated by DeepAM, is used as the seed schedule for DeepCAM. The household head is selected based on a priority ranking that considers age, employment status, possession of a driver license, vehicle access, and gender. 

\begin{figure}
  \centering
  \includegraphics[width=0.35\textwidth]{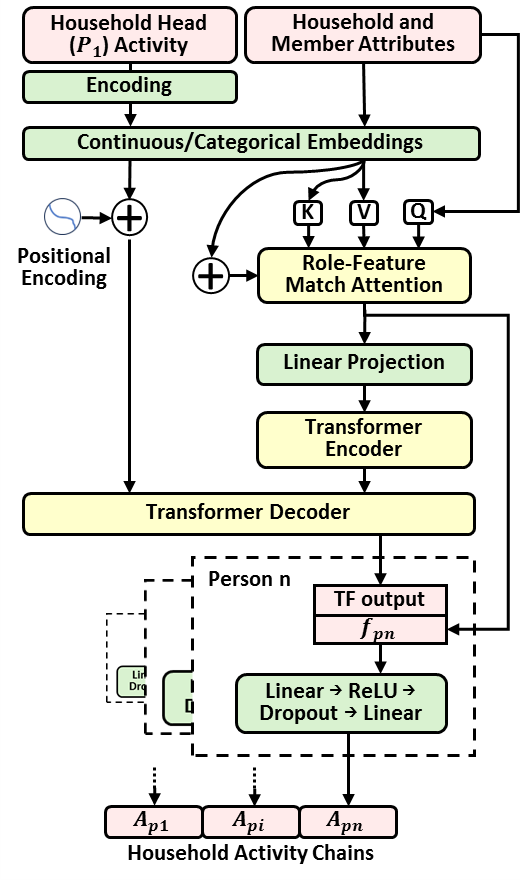}
  \caption{Network structure of DeepCAM.}
  \label{fig:network}
\end{figure}

\subsubsection{Household Coordinated Activity Generation}

The proposed DeepCAM model is a role-aware multi-person activity generation network designed to capture both individual-specific behaviors and coordinated group dynamics, as shown in Fig. \ref{fig:network}, taking two inputs, i.e., an activity chain and a set of person features.

The modeling process begins by encoding the one-day activity chains $A_{p_i}$, where each day is divided into 96 time slots, each representing a 15-minute interval and labeled with the corresponding activity type. This encoding serves as the temporal input signal for the network. 

To support generating activities for specific individuals, the model uses a \textit{role-feature matching attention mechanism} that processes household $h$ and member attributes $f$. While a standard transformer~\cite{vaswani2017attention} allows sharing information among household members—critical for modeling coordination—it can blur individual identities, particularly in households with similar members, such as siblings.

To address this, the role-feature match attention layer introduces explicit role-person associations. It takes learned query vectors representing role positions and aligns them with the most relevant individual attributes using a soft matching strategy. This introduces a diagonal bias into the attention map, encouraging members to associate with their attributes while retaining flexibility for less clearly defined roles. This helps the model identify “who is who” before making any behavioral predictions, ensuring that the predicted activities are correctly linked to each individual rather than being arbitrarily assigned. A residual connection from original attribute embeddings preserves individual identity throughout.

The refined person features pass through a transformer encoder to capture interdependencies among members. A decoder then integrates these with the input activity sequence, conditioning predictions on both person context and prior activities. Each decoder output is concatenated with individual features and fed through fully connected layers to produce activity logits. A masking mechanism handles variable household sizes by restricting predictions to valid individuals. This architecture enables DeepCAM to model coordinated household behavior by embedding roles and identities at multiple stages.

\subsubsection{Loss Function Design}

The overall training loss consists of a cross-entropy term and an auxiliary regularization term designed to penalize overconfident activity predictions:

\[
L_{\text{total}} = L_{\text{CE}} + \lambda_{\text{AOR}} \cdot (R_{\text{individual}} + R_{\text{household}})
\]

where $\lambda_{\text{AOR}}$ controls the weight of the regularization term. The cross-entropy loss is defined as: $L_{\text{CE}} = - \sum_{i} a_i \log(\hat{a}_i)$, which encourages the model to assign high probability to the correct activity label. However, it does not penalize the model for assigning high probabilities to multiple conflicting activities, potentially leading to overconfident or ambiguous predictions.

To address this, we introduce \textit{Activity Overconfidence Regularization}, which includes two components. The first, $R_{\text{individual}}$, penalizes per-person overconfidence by measuring the total predicted probability assigned to incorrect activities:

\[
R_{\text{individual}} =
\frac{
\sum\limits_{b,t,p,a} \max(0, \hat{p}_{b,t,p,a} - y_{b,t,p,a}) \cdot w_a
}{
N_{\text{individual}}
}
\]

The second, $R_{\text{household}}$, penalizes the overestimation of group sizes by comparing the predicted and actual total number of participants for each activity:

\[
R_{\text{household}} =
\frac{
\sum\limits_{b,t,a} \max\left(0, \sum_p \hat{p}_{b,t,p,a} - \sum_p y_{b,t,p,a} \right) \cdot w_a
}{
N_{\text{household}}
}
\]

Here, $b$, $t$, $p$, and $a$ denote indices over batch samples, time steps, household members, and activity types, respectively. The variable $\hat{p}_{b,t,p,a}$ represents the softmax probability that person $p$ at time $t$ in batch $b$ is assigned to activity $a$, and $y_{b,t,p,a} \in \{0,1\}$ is the corresponding one-hot encoded ground truth. $N_{\text{individual}}$ and $N_{\text{household}}$ denote the total number of person-activity and household-level entries, respectively. The activity weights $w_a$ reflect the degree to which each activity is expected to be performed individually, derived from NHTS.

\subsubsection{Event Table as Coordinated Activity Representation}
To support structured representation for travel demand modeling and simulation, an event table is created to record all activities, assigning each activity an event ID. This allows multiple household members to share the same event ID when participating in the same activity. Although this paper focuses on household-level events, the structure provides a foundation for scaling to larger gatherings involving individuals from different households.

Coordinated activities within households are identified by grouping activities involving multiple household members that start within a 15-minute window, following the approach in \cite{bhat2012household, kang2008integrated}. These events are further examined to determine coordination: if all involved individuals report the same activity type, or if one or more report an "accompanying" type (e.g., escort), the event is labeled as a coordinated activity. 

\subsection{Activity Location Assignment}
To spatially ground the generated activity chains, we adapt a previously developed activity location assignment module~\cite{liao2024deep}, which maps each activity to a TAZ. The algorithm follows three steps:

\subsubsection{Mandatory Activities} Home, work, and school locations are assigned first. Commute distances are sampled from regional distributions, and TAZs are selected by minimizing deviation from these targets while ensuring land use compatibility:   $Z_{md} = \mathop{\arg\min}_{z \in \mathcal{Z}} \left\{ d(h,z) - \hat{d} \right\}$, where \( h \) is the home TAZ, \( z \in \mathcal{Z} \) is a candidate TAZ, \( \hat{d} \) is the sampled commute distance, and \( d(h, z) \) is the network distance between \( h \) and \( z \).

\subsubsection{Non-Mandatory Activities} Discretionary activities are placed between anchors by minimizing a weighted cost of travel distance and directional deviation, while respecting travel time limits: $Z_{nmd} = \mathop{\arg\min}_{z \in \mathcal{Z}} \left\{ \alpha \left| d - \hat{d} \right| + \beta \left| \theta - \hat{\theta} \right| \right\}, \quad t(z_i, z_{i+1}) \leq T_{max}$. Here, \( z_{prev} \) and \( z_{next} \) are the previous and next anchor TAZs, \( \hat{d} \) and \( \hat{\theta} \) are sampled distance and angular deviation values, \( \alpha \) and \( \beta \) are weighting factors, and \( t(\cdot) \) is the travel time function constrained by the available time window \( T_{\max} \).

\subsubsection{Spatial Refinement} To match observed spatial distributions $D$, the activity frequencies are iteratively adjusted: $D^{t+1} = D^t + \eta \cdot (F_{\text{target}} - F_{\text{current}})$, where \( F_{\text{current}} \) and \( F_{\text{target}} \) are the current and reference spatial distributions, and \( \eta \) is a learning rate.

\begin{figure}
  \centering
  \includegraphics[width=0.45\textwidth]{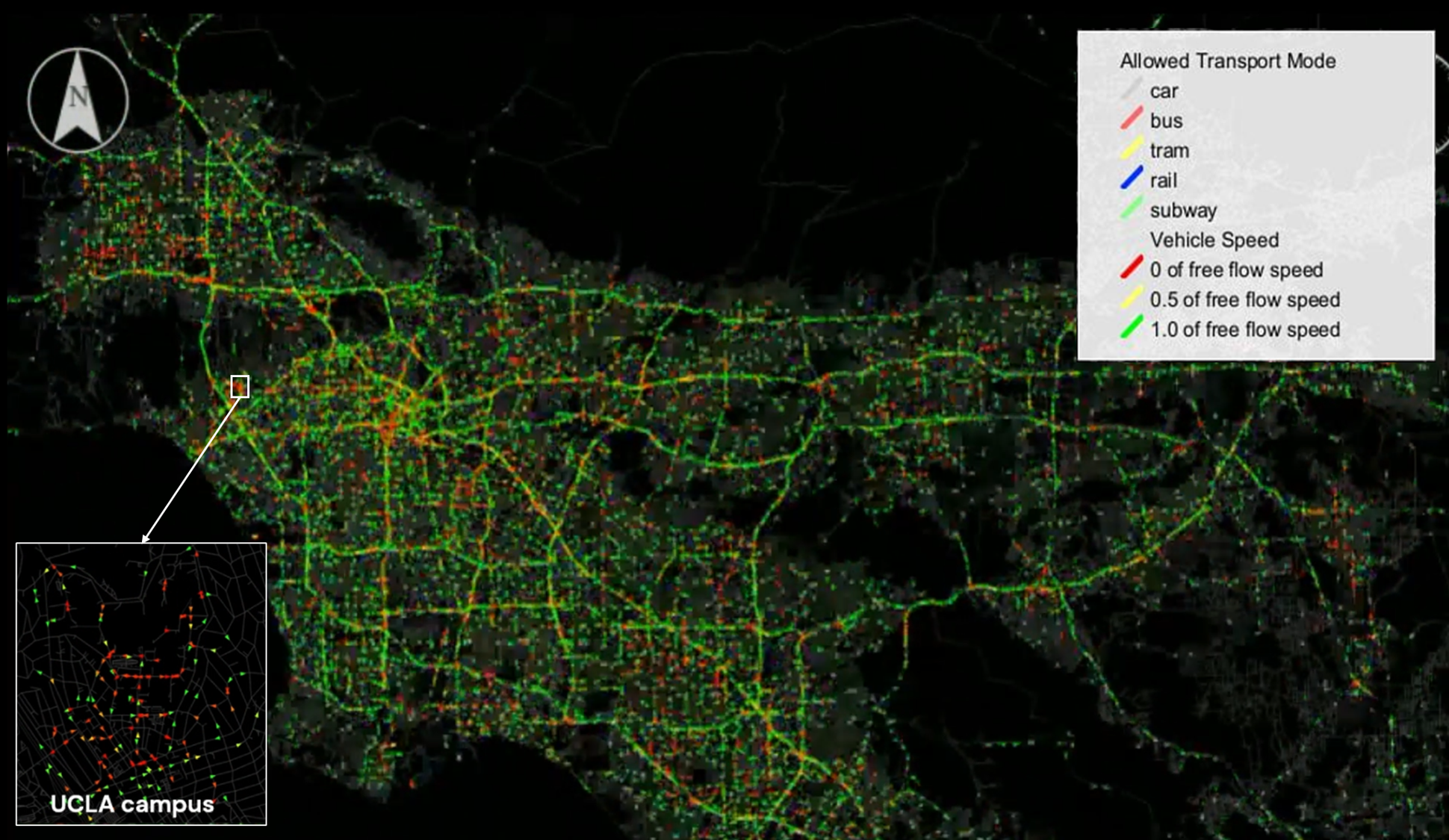}
  \caption{Traffic simulation for LA using MATSim.}
  \label{fig:simulation}
\end{figure}

\subsection{Traffic Simulation}
In addition to activity patterns and origin-destination (TAZ pair) data, this study uses traffic simulation to generate fine-grained trajectories, as presented in Fig.~\ref{fig:simulation}. By incorporating real-world traffic conditions, network dynamics, and agent interactions, we add another layer to travel diaries with route choice, mode choice, and second-by-second trajectories, describing where, when, how, and why people travel. Moreover, simulation provides an additional validation dimension by comparing generated traffic with observed data.

\subsubsection{Mode Choice}
Simulating traffic allows assigning realistic mode choices to each trip, so MATSim is adopted as the simulation platform \cite{horni2016multi} for mode assignment. Based on household car ownership and regional mode share statistics \cite{laalmanaccommuting2023}, we first initialize the mode assignments, and then MATSim iteratively adjusts these mode choices through its co-evolutionary algorithm, allowing agents to adapt their travel behavior based on experienced utility. The mode choice utility functions have been carefully calibrated as part of the LASim framework to accurately reflect travel preferences in the LA region.

\subsubsection{Simulation Setup}
The simulation framework integrates several components. The synthetic population is first assigned with daily schedules specifying activities, timings, locations, and initial travel modes. The high-resolution multimodal transportation network is constructed from OpenStreetMap, including calibrated attributes such as link capacities and lane numbers. Public transit operations, based on GTFS (General Transit Feed Specification) data \cite{lametrogtfs2016}, are incorporated to support realistic multimodal trip chains. The simulation leverages the calibrated LASim model \cite{heABMtrans}, with utility parameters for travel time, monetary cost, transfers, and waiting time optimized to match observed travel behavior in LA. Traffic flow dynamics are also calibrated using observed count data from Caltrans PeMS \cite{pems2020}.

Through this simulation setup, we can evaluate not only the statistical properties of our generated activity patterns but also their emergent traffic impacts when executed within a realistic transportation environment.

\section{Experiments}
\subsection{Data and Experimental Setup}
The 2017 National Household Travel Survey (NHTS)~\cite{nhtsdata} is used for training, which includes demographics, activity patterns, and travel behaviors from over 129,600 U.S. households. Given the standardized structure of household travel surveys, the methodology proposed in this work can be readily applied to other regions. The original 19 activity types are aggregated into 15 types based on their associated locations, and the data is split into training, validation, and test sets in a 0.8:0.1:0.1 ratio.

To enable a fair comparison with legacy ABM pipelines, we adopt the same synthetic population as used in the SCAG ABM~\cite{scag2020} generated by the SimAGENT system~\cite{pendyala2012simagent}. Both our framework and the legacy ABM are applied to simulate a population of one million residents in the LA region. The outputs of both models are evaluated against observed real-world traffic data from Caltrans Performance Measurement System (PeMS) \cite{pems2020} in the LA region, which serves as the ground truth for validation. 

The models are first trained on NHTS and then transferred to LA region for large-scale simulation. Prior work \cite{liao2024deep} has shown the model’s transferability using limited local data~\cite{calidata}. Training DeepAM on NHTS takes 3 hours on an NVIDIA A6000 GPU, while DeepCAM trains in 35 minutes. DeepAM generates activity schedules for 386,091 SCAG household heads in 10 minutes, and DeepCAM completes inference for one million individuals in 35 seconds. These results highlight the framework’s efficiency and scalability for large-scale population synthesis.

\subsection{Evaluation Metrics and Results}
We evaluate the proposed framework by assessing the quality of outputs generated by DeepCAM and the realism of the proposed travel demand framework. 

\subsubsection{Activity Chain Quality}
While the initial DeepAM model for generating individual activity chains was validated in previous study \cite{liao2024deep}, here we assess DeepCAM’s ability to generate plausible schedules for all household members. We evaluate distributional similarity via Jensen-Shannon Divergence (JSD) in activity types, timing, and activity counts.

\begin{figure}
  \centering
  \begin{subfigure}[b]{0.48\textwidth}
    \includegraphics[width=\textwidth]{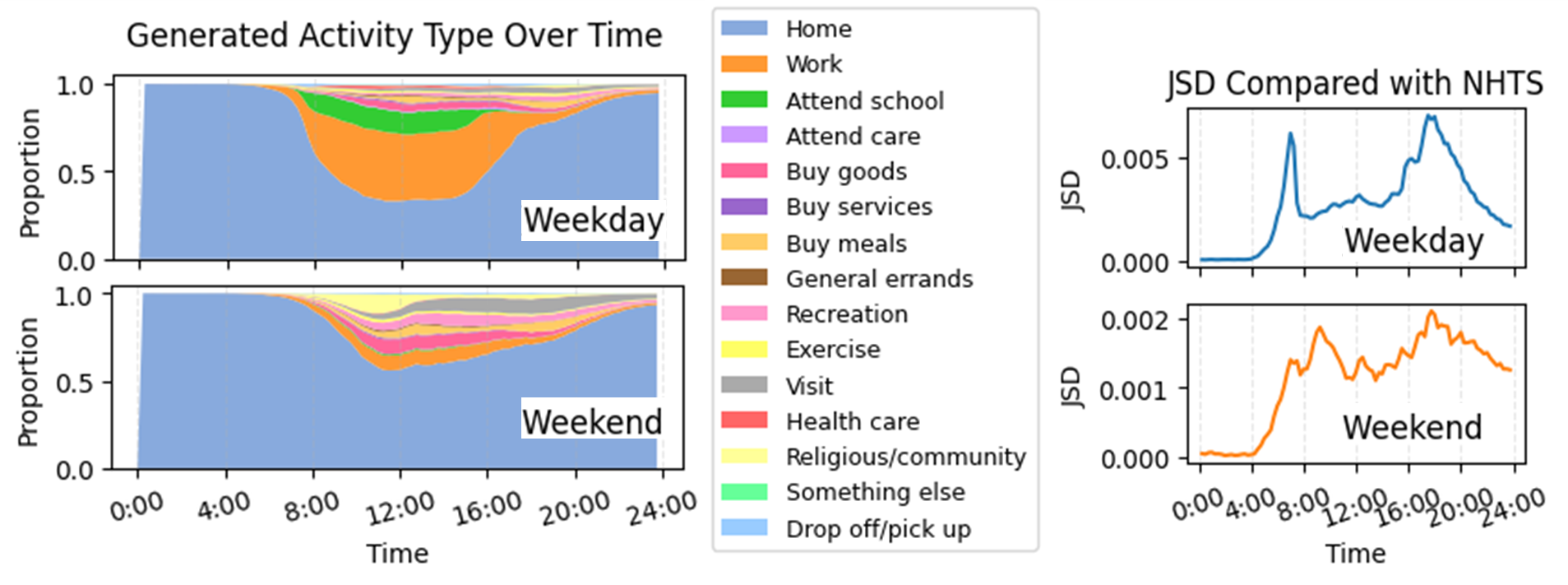}
    \caption{Activity pattern over one day and similarity (JSD) with NHTS}
    \label{fig:pattern}
  \end{subfigure}
  \begin{subfigure}[b]{0.48\textwidth}
    \includegraphics[width=\textwidth]{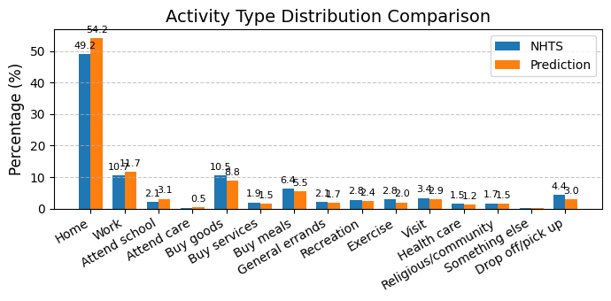}
    \caption{Aggregated activity type distribution}
    \label{fig:actdis}
  \end{subfigure}
  \begin{subfigure}[b]{0.48\textwidth}
    \includegraphics[width=\textwidth]{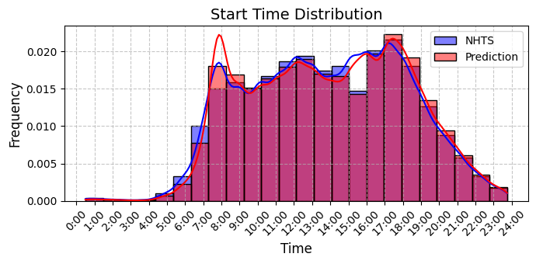}
    \caption{Aggregated start time distribution}
    \label{fig:starttime}
  \end{subfigure}
  \begin{subfigure}[b]{0.48\textwidth}
    \includegraphics[width=\textwidth]{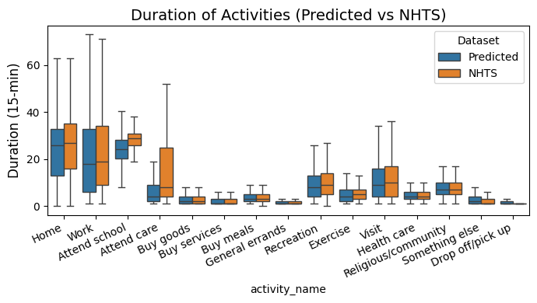}
    \caption{Aggregated duration}
    \label{fig:duration}
  \end{subfigure}
  \begin{subfigure}[b]{0.48\textwidth}
    \includegraphics[width=\textwidth]{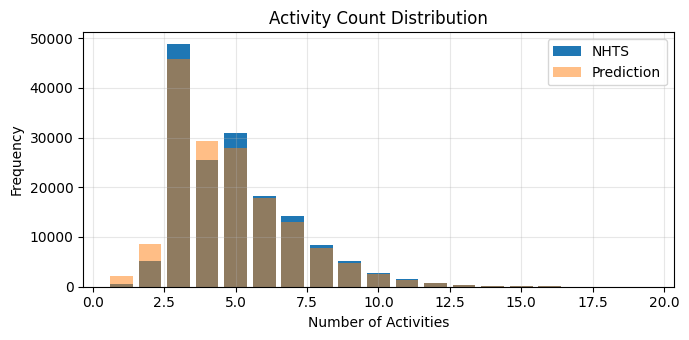}
    \caption{Number of daily activities}
    \label{fig:length}
  \end{subfigure}
  \caption{Daily activity patterns: model prediction vs. NHTS.}
  \label{fig:daily_patterns}
\end{figure}

The quality of the generated activity chains is assessed as in Fig.~\ref{fig:daily_patterns}. The daily activity distributions over time for both weekdays and weekends are presented in Fig.~\ref{fig:daily_patterns}(a) showing clear differences: work and school activities are more frequent on weekdays, while religious and recreational activities are more common on weekends. The corresponding JSD curves remain consistently low, with a maximum of 0.006, indicating strong agreement between predicted schedules and NHTS data in temporal activity allocation. The aggregated distribution of activity types is shown in Fig.~\ref{fig:daily_patterns}(b). The predicted distribution closely matches the observed data across all activity categories, with a JSD of 0.005. The start time distributions across all activity instances is illustrated in Fig.~\ref{fig:daily_patterns}(c). The peak activity initiation times in both the morning and late afternoon are well captured by the model. A very low JSD of 0.002 is reported, indicating strong temporal alignment of activities. Figure~\ref{fig:daily_patterns}(d) compares predicted and observed activity durations by type, showing close alignment in medians and variability for most activities. Minor discrepancies appear in sparse categories like attend care, which represents only 0.3\% of all activities. Overall, duration patterns are reproduced with reasonable accuracy. Figure~\ref{fig:daily_patterns}(e) evaluates the number of activities per agent per day. The predicted distribution closely matches the NHTS data, with a JSD of 0.005. Most individuals perform 2 to 6 activities daily, peaking at 3, which primarily reflects "home–work–home" schedules. 
\begin{figure}
  \centering
  \includegraphics[width=0.49\textwidth]{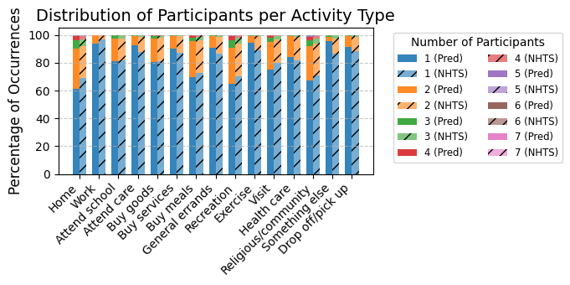}
  \caption{Household activity participation per activity type.}
  \label{fig:participant}
\end{figure}

\begin{figure*}[h]
  \centering
  \includegraphics[width=0.99\textwidth]{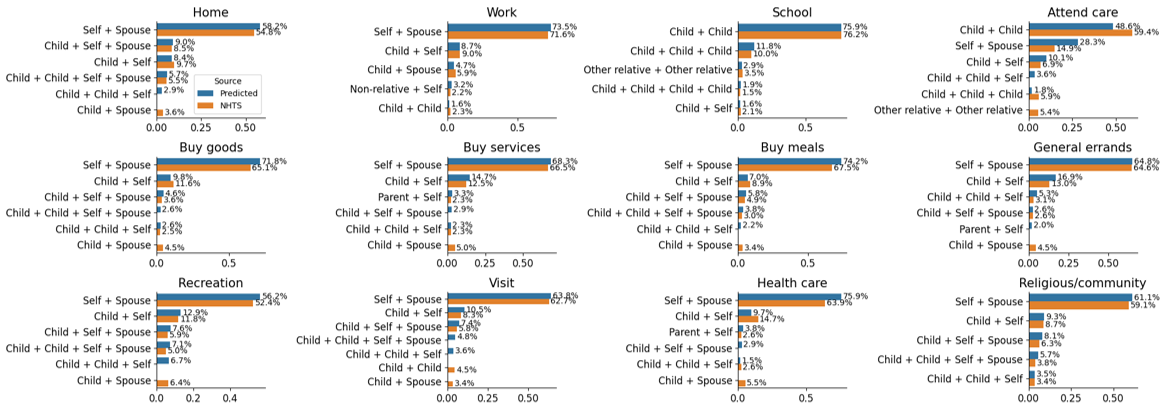}
  \caption{Most frequent role combinations per activity type.}
  \label{fig:role}
\end{figure*}

\subsubsection{Coordinated Activity Evaluation}
Besides generating realistic activity chains for individual, DeepCAM’s ability to model intra-household coordination needs to be evaluated. We examine the distribution of participants across activity types to assess joint participation behavior, and compare the demographic composition of participants in joint activities with patterns observed in the NHTS, such as parents with children or spouses participating together.

The distribution of participants per activity type is elaborated in Fig.~\ref{fig:participant}. Most activities are performed individually, with over 85\% involving a single participant across both the predicted data and NHTS, aligning with the finding in \cite{gliebe2005modeling}. Joint participation in activities such as buy goods, recreation, meals, and religious/community events is also captured, with distributions closely matching the empirical data. The alignment indicates that DeepCAM is able to replicate observed coordination patterns within households, particularly for shared discretionary and family-oriented activities.

DeepCAM learns not only how many people coordinate the activity but also captures realistic co-participation structures across household roles. As illustrated in Fig. ~\ref{fig:role}, the most frequent role combinations per activity type are shown for both the predicted data and the NHTS. Key coordination patterns are well reproduced. For example, self + spouse is consistently shown as the dominant pairing for buy goods, buy services, general errands, and recreation, reflecting accurate modeling of spousal routines. Child + child combinations dominate school, while religious/community and visit activities frequently involve self + spouse and mixed-generation groups such as child + self or parent + self. Less common combinations, including non-relative (e.g., housemates) + self, are also captured for certain discretionary activities like work and visit. Overall, the predicted role combinations align well with observed patterns, though small discrepancies are observed in low-frequency activities such as attend care, where the model tends to overrepresent child-dominant groups.

\subsubsection{Full-Pipeline Simulation Validation}
To validate the complete framework, we assess the realism of aggregate travel demand after location assignment and traffic simulation. Simulated traffic volume, speed, and flow are compared against real-world observations from the PeMS, which serves as the ground truth. In addition to absolute comparison, we benchmark our results against outputs from the legacy SCAG ABM, which represents the baseline for current regional modeling practice. Aggregate measures such as Vehicle Miles Traveled (VMT), origin-destination (OD) matrices, traffic speed, and traffic volume are evaluated.

\begin{figure}[ht]
  \centering
  \includegraphics[width=0.49\textwidth]{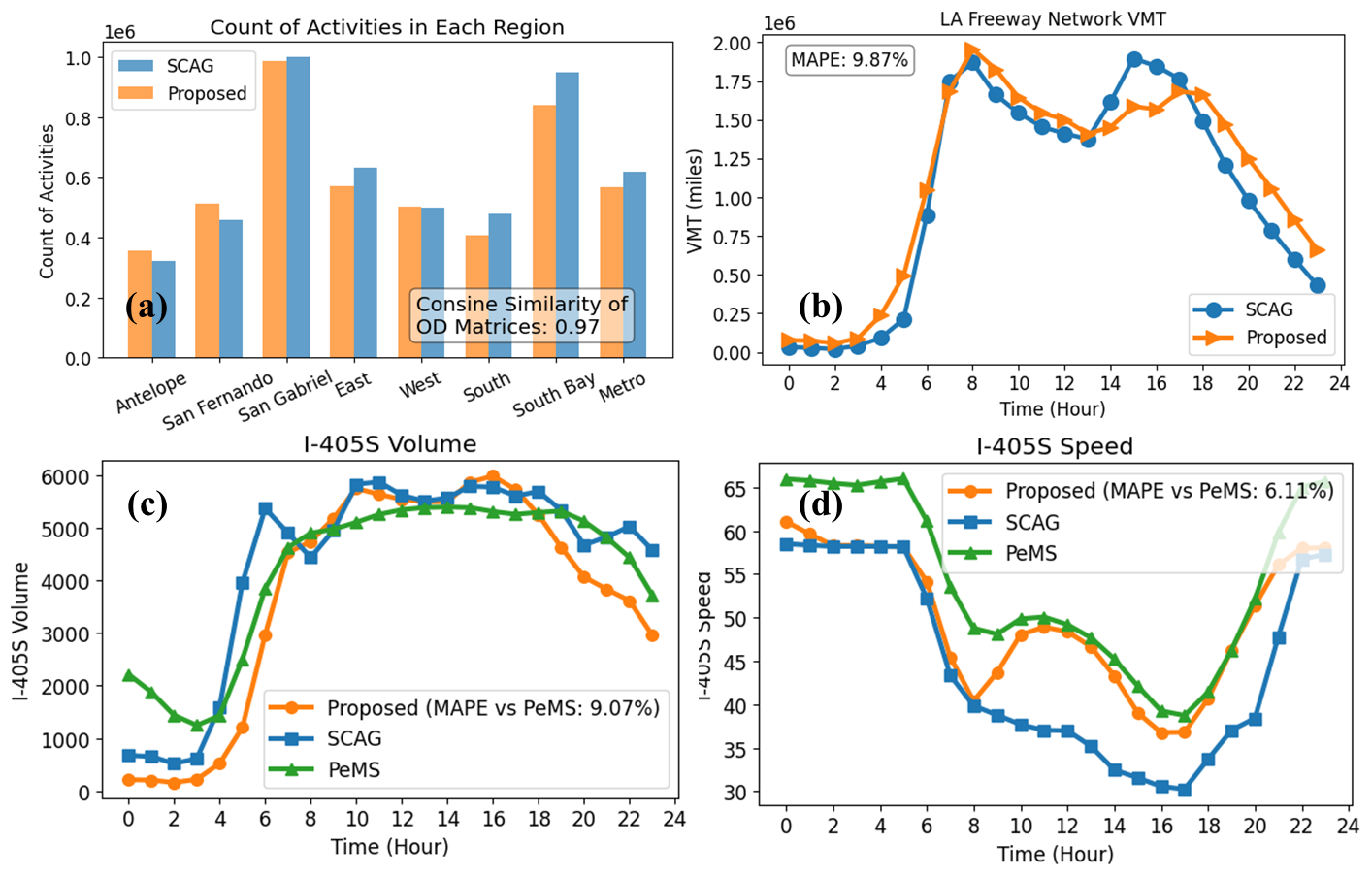}
  \caption{Validation for activity location assignment and traffic loading at network level and corridor level}
  \label{fig:simulationRT}
\end{figure}

As shown in Fig.~\ref{fig:simulationRT}(a), the assigned activity locations closely match the result of SCAG ABM across all LA sub-regions, with a cosine similarity of 0.97 between OD matrices. Fig.~\ref{fig:simulationRT}(b) presents the simulated freeway-level VMT over 24 hours. The proposed framework closely tracks SCAG ABM patterns, achieving a JSD of 0.006 and a MAPE of 9.8\%, closely replicating SCAG ABM’s system-level traffic dynamics while offering significantly lower modeling cost and improved computational efficiency.

The corridor-level comparison is presented in Fig.~\ref{fig:simulationRT}(c) and (d), where the proposed method is compared with SCAG and PeMS data, revealing high traffic demand and recurring congestion during daytime hours in the southbound direction of the studied I-405 corridor, covered by 20 loop detectors, around 8.5 miles. Congestion on the southbound direction emerges primarily after noon and persists into the evening peak. The proposed framework achieves a MAPE of 9.07\% for traffic volume and 6.11\% for speed, and a JSD of 0.022 for traffic volume and 0.001 for speed, indicating strong consistency with observed traffic dynamics across both flow and speed dimensions. During the light traffic hours (0:00–5:00 and 22:00-24:00), the proposed framework slightly underestimates free-flow speeds compared to PeMS observations, as shown in Fig.~\ref{fig:simulationRT}(d). This bias is partially due to the preset speed limit in MATSim, leading to reduced early-morning free-flow speeds. 

\section{Conclusion and Future Work}
This paper presents a travel demand modeling framework that extends deep generative activity models to capture intra-household coordination. The modular pipeline integrates population synthesis, activity generation (individual and coordinated), spatial assignment, and traffic simulation. We demonstrate the model’s ability to generate realistic daily activity patterns at both individual and household levels and validate the resulting demand using observed traffic data from a large-scale LA case study. The model effectively captures temporal, spatial, and social dynamics while remaining efficient and transferable.

Future work will extend coordination modeling to social networks beyond households, such as friends and co-workers. We also plan to incorporate real-world mobility data, like GPS traces, to complement survey-based inputs and address limitations such as sampling bias and self-reporting errors.

\section*{Acknowledgment}
Supported by the Intelligence Advanced Research Projects Activity (IARPA) via the Department of Interior/Interior Business Center (DOI/IBC) contract number 140D0423C0033. The U.S. Government is authorized to reproduce and distribute reprints for Governmental purposes notwithstanding any copyright annotation thereon. Disclaimer: The views and conclusions contained herein are those of the authors and should not be interpreted as necessarily representing the official policies or endorsements, either expressed or implied, of IARPA, DOI/IBC, or the U.S. Government.

\bibliographystyle{IEEEtran}
\bibliography{reference}

\begin{thebibliography}{10}
\providecommand{\url}[1]{#1}
\csname url@samestyle\endcsname
\providecommand{\newblock}{\relax}
\providecommand{\bibinfo}[2]{#2}
\providecommand{\BIBentrySTDinterwordspacing}{\spaceskip=0pt\relax}
\providecommand{\BIBentryALTinterwordstretchfactor}{4}
\providecommand{\BIBentryALTinterwordspacing}{\spaceskip=\fontdimen2\font plus
\BIBentryALTinterwordstretchfactor\fontdimen3\font minus \fontdimen4\font\relax}
\providecommand{\BIBforeignlanguage}[2]{{%
\expandafter\ifx\csname l@#1\endcsname\relax
\typeout{** WARNING: IEEEtran.bst: No hyphenation pattern has been}%
\typeout{** loaded for the language `#1'. Using the pattern for}%
\typeout{** the default language instead.}%
\else
\language=\csname l@#1\endcsname
\fi
#2}}
\providecommand{\BIBdecl}{\relax}
\BIBdecl

\bibitem{ma2024mobility}
H.~Ma, Y.~Liu, Q.~Jiang, B.~Y. He, X.~Liao, and J.~Ma, ``Mobility ai agents and networks,'' \emph{IEEE Transactions on Intelligent Vehicles}, 2024.

\bibitem{pappalardo2016analytical}
L.~Pappalardo, M.~Vanhoof, L.~Gabrielli, Z.~Smoreda, D.~Pedreschi, and F.~Giannotti, ``An analytical framework to nowcast well-being using mobile phone data,'' \emph{International Journal of Data Science and Analytics}, vol.~2, pp. 75--92, 2016.

\bibitem{abidi2023mobility}
J.~Abidi and F.~Filali, ``Mobility analytics of fans during the 2021 fifa arab cup tm football tournament in qatar,'' \emph{IEEE Open Journal of Intelligent Transportation Systems}, 2023.

\bibitem{bohm2021quantifying}
M.~Bohm, M.~Nanni, and L.~Pappalardo, ``Quantifying the presence of air pollutants over a road network in high spatio-temporal resolution,'' in \emph{Climate Change AI, NeurIPS Workshop}, 2021.

\bibitem{ben1985discrete}
M.~E. Ben-Akiva and S.~R. Lerman, \emph{Discrete choice analysis: theory and application to travel demand}.\hskip 1em plus 0.5em minus 0.4em\relax MIT press, 1985, vol.~9.

\bibitem{horowitz1980utility}
J.~Horowitz, ``A utility maximizing model of the demand for multi-destination non-work travel,'' \emph{Transportation Research Part B: Methodological}, vol.~14, no.~4, pp. 369--386, 1980.

\bibitem{algers1995stockholm}
S.~Algers, A.~Daly, P.~Kjellman, and S.~Widlert, ``Stockholm model system (sims): Application,'' in \emph{7th World Conference of Transportation Research}.\hskip 1em plus 0.5em minus 0.4em\relax Sydney Australia, 1995, pp. 16--21.

\bibitem{rossi1997tour}
T.~F. Rossi and Y.~Shiftan, ``Tour based travel demand modeling in the us,'' \emph{IFAC Proceedings Volumes}, vol.~30, no.~8, pp. 381--386, 1997.

\bibitem{goulias2011simulator}
K.~G. Goulias, C.~R. Bhat, R.~M. Pendyala, Y.~Chen, R.~Paleti, K.~C. Konduri, G.~Huang, and H.-H. Hu, ``Simulator of activities, greenhouse emissions, networks, and travel (simagent) in southern california: Design, implementation, preliminary findings, and integration plans,'' in \emph{2011 IEEE Forum on Integrated and Sustainable Transportation Systems}.\hskip 1em plus 0.5em minus 0.4em\relax IEEE, 2011, pp. 164--169.

\bibitem{bowman2001activity}
J.~L. Bowman and M.~E. Ben-Akiva, ``Activity-based disaggregate travel demand model system with activity schedules,'' \emph{Transportation research part a: policy and practice}, vol.~35, no.~1, pp. 1--28, 2001.

\bibitem{bhat2012household}
C.~R. Bhat, K.~G. Goulias, R.~M. Pendyala, R.~Paleti, R.~Sidharthan, L.~Schmitt, and H.-h. Hu, ``A household-level activity pattern generation model for the simulator of activities, greenhouse emissions, networks, and travel (simagent) system in southern california,'' in \emph{91st Annual Meeting of the Transportation Research Board, Washington, DC}, 2012.

\bibitem{liao2024deep}
X.~Liao, Q.~Jiang, B.~Y. He, Y.~Liu, C.~Kuai, and J.~Ma, ``Deep activity model: A generative approach for human mobility pattern synthesis,'' \emph{arXiv preprint arXiv:2405.17468}, 2024.

\bibitem{adnan2016simmobility}
M.~Adnan, F.~C. Pereira, C.~M.~L. Azevedo, K.~Basak, M.~Lovric, S.~Raveau, Y.~Zhu, J.~Ferreira, C.~Zegras, and M.~Ben-Akiva, ``Simmobility: A multi-scale integrated agent-based simulation platform,'' in \emph{95th Annual Meeting of the Transportation Research Board Forthcoming in Transportation Research Record}, vol.~2.\hskip 1em plus 0.5em minus 0.4em\relax The National Academies of Sciences, Engineering, and Medicine Washington, DC, 2016.

\bibitem{freedman2023activitysim}
J.~Freedman and D.~Hensle, ``Activitysim: Activity-based travel demand modeling built by and for users (rsg, 2021),'' 2021.

\bibitem{shone2025modelling}
F.~Shone and T.~Hillel, ``Modelling activity scheduling behaviour with deep generative machine learning,'' \emph{arXiv preprint arXiv:2501.10221}, 2025.

\bibitem{gliebe2005modeling}
J.~P. Gliebe and F.~S. Koppelman, ``Modeling household activity--travel interactions as parallel constrained choices,'' \emph{Transportation}, vol.~32, pp. 449--471, 2005.

\bibitem{bradley2005model}
M.~Bradley and P.~Vovsha, ``A model for joint choice of daily activity pattern types of household members,'' \emph{Transportation}, vol.~32, pp. 545--571, 2005.

\bibitem{zhang2009modeling}
J.~Zhang, M.~Kuwano, B.~Lee, and A.~Fujiwara, ``Modeling household discrete choice behavior incorporating heterogeneous group decision-making mechanisms,'' \emph{Transportation Research Part B: Methodological}, vol.~43, no.~2, pp. 230--250, 2009.

\bibitem{shakeel2022joint}
F.~Shakeel, M.~Adnan, and T.~Bellemans, ``Joint-activities generation among household members using a latent class model,'' \emph{Transportation Research Procedia}, vol.~62, pp. 557--564, 2022.

\bibitem{arentze2009need}
T.~A. Arentze and H.~J. Timmermans, ``A need-based model of multi-day, multi-person activity generation,'' \emph{Transportation Research Part B: Methodological}, vol.~43, no.~2, pp. 251--265, 2009.

\bibitem{bhat2013household}
C.~R. Bhat, K.~G. Goulias, R.~M. Pendyala, R.~Paleti, R.~Sidharthan, L.~Schmitt, and H.-H. Hu, ``A household-level activity pattern generation model with an application for southern california,'' \emph{Transportation}, vol.~40, pp. 1063--1086, 2013.

\bibitem{mcnally2014popgen}
K.~McNally, R.~Cotton, A.~Hogg, and G.~Loizou, ``Popgen: a virtual human population generator,'' \emph{Toxicology}, vol. 315, pp. 70--85, 2014.

\bibitem{marg2016popgen}
{Mobility Analytics Research Group}, ``{PopGen: Synthetic Population Generator},'' \url{http://www.mobilityanalytics.org/popgen.html}, 2016, accessed April 21, 2025.

\bibitem{pendyala2012simagent}
R.~M. Pendyala, C.~R. Bhat, K.~G. Goulias, R.~Paleti, K.~Konduri, R.~Sidharthan, and K.~P. Christian, ``Simagent population synthesis,'' \emph{Synthesis}, 2012.

\bibitem{farooq2013simulation}
B.~Farooq, M.~Bierlaire, R.~Hurtubia, and G.~Fl{\"o}tter{\"o}d, ``Simulation based population synthesis,'' \emph{Transportation Research Part B: Methodological}, vol.~58, pp. 243--263, 2013.

\bibitem{vaswani2017attention}
A.~Vaswani, N.~Shazeer, N.~Parmar, J.~Uszkoreit, L.~Jones, A.~N. Gomez, {\L}.~Kaiser, and I.~Polosukhin, ``Attention is all you need,'' \emph{Advances in neural information processing systems}, vol.~30, 2017.

\bibitem{kang2008integrated}
H.~Kang and D.~M. Scott, ``An integrated spatio-temporal gis toolkit for exploring intra-household interactions,'' \emph{Transportation}, vol.~35, pp. 253--268, 2008.

\bibitem{horni2016multi}
A.~Horni, K.~Nagel, and K.~Axhausen, Eds., \emph{Multi-Agent Transport Simulation MATSim}.\hskip 1em plus 0.5em minus 0.4em\relax London: Ubiquity Press, Aug 2016.

\bibitem{laalmanaccommuting2023}
\BIBentryALTinterwordspacing
{Los Angeles Almanac}, ``Commuting to work in los angeles county,'' 2023, accessed: 2025-04-28. [Online]. Available: \url{https://laalmanac.com/employment/em22.php}
\BIBentrySTDinterwordspacing

\bibitem{lametrogtfs2016}
\BIBentryALTinterwordspacing
{Los Angeles County Metropolitan Transportation Authority}, ``La metro rail gtfs,'' 2016, accessed: 2025-04-28. [Online]. Available: \url{https://transitfeeds.com/p/la-metro/677?p=176}
\BIBentrySTDinterwordspacing

\bibitem{heABMtrans}
B.~Y. He, Q.~Jiang, H.~Ma, and J.~Ma, ``Multi-agent multimodal transportation simulation for mega-cities: Application of los angeles,'' 2024.

\bibitem{pems2020}
\BIBentryALTinterwordspacing
{California Department of Transportation}, ``{PeMS},'' 2020, retrieved from California Department of Transportation. [Online]. Available: \url{https://pems.dot.ca.gov}
\BIBentrySTDinterwordspacing

\bibitem{nhtsdata}
F.~H. Administration, ``National household travel survey,'' \url{https://nhts.ornl.gov}, 2017.

\bibitem{scag2020}
{Southern California Association of Governments (SCAG)}, ``{2016 Regional Travel Demand Model and Model Validation Report},'' {Southern California Association of Governments}, Tech. Rep., 2020.

\bibitem{calidata}
C.~D. of~Transportation, ``California household travel survey,'' \url{https://www.nrel.gov/transportation/secure-transportation-data/tsdc-california-travel-survey.html}, 2012.

\end{thebibliography}

\end{document}